# Mitigation of Spatial Nonstationarity with Vision Transformers

Lei Liu[a], Javier E. Santos[b], Maša Prodanović[a], and Michael J. Pyrcz[a,c]

[a]Hildebrand Department of Petroleum and Geosystems Engineering, The University of Texas at Austin, Austin, Texas, USA

[b]Center for NonLinear Studies, Los Alamos National Laboratory, Los Alamos, New Mexico, USA

[c]Department of Geological Sciences, The University of Texas at Austin, Austin, Texas, USA

Authorship contribution statement

Lei Liu: Conceptualization, Methodology, Software, Writing - Original Draft. Javier E. Santos: Conceptualization, Validation, Writing - Review & Editing. Maša Prodanović: Writing - Review & Editing. Michael J. Pyrcz: Conceptualization, Methodology, Writing - Review & Editing, Supervision, Funding acquisition.

ABSTRACT

Spatial nonstationarity, the location variance of features' statistical distributions, is ubiquitous in many natural settings. For example, in geological reservoirs rock matrix porosity varies vertically due to geomechanical compaction trends, in mineral deposits grades vary due to sedimentation and concentration processes, in hydrology rainfall varies due to the atmosphere and topography interactions, and in metallurgy crystalline structures vary due to differential cooling. Conventional geostatistical modeling workflows rely on the assumption of stationarity to be able to model spatial features for the geostatistical inference. Nevertheless, this is often not a realistic assumption when dealing with nonstationary spatial data and this has motivated a variety of nonstationary spatial modeling workflows such as trend and residual decomposition, cosimulation with secondary features, and spatial segmentation and independent modeling over stationary subdomains. The advent of deep learning technologies has enabled new workflows for modeling spatial relationships. However, there is a paucity of demonstrated best practice and general guidance on mitigation of spatial nonstationarity with deep learning in the geospatial context. We demonstrate the impact of two common types of geostatistical spatial nonstationarity on deep learning model prediction performance and propose the mitigation of such impacts using self-attention (vision transformer) models. We demonstrate the utility of vision transformers for the mitigation of nonstationarity with relative errors as low as 10%, exceeding the performance of alternative deep learning methods such as convolutional neural networks. We establish best practice by demonstrating the ability of self-attention networks for modeling large-scale spatial relationships in the presence of commonly observed geospatial nonstationarity.

1. Introduction

Spatial stationarity, defined as the statistics of interest are invariant over the spatial or temporal interval of interest, is a fundamental decision required by many statistical and geostatistical modeling methods (Pyrcz and Deutsch, 2014; Ripley, 2004). However, the decision of stationarity should be made judiciously because spatial nonstationarity is ubiquitous in subsurface settings due to spatiotemporal changes in the fundamental processes responsible for sediment supply, deposition, preservation, and alteration, e.g., sequence stratigraphic processes such as regressive and transgressive systems tracks produced during eustatic change (Prokoph and Barthelmes, 1996), nonstationary hydrologic behavior due to the atmosphere and topography interactions (Mallick et al., 2018), geomechanical compaction trends or sedimentation records in geological reservoirs (van Thienen-Visser and Breunese, 2015). Geostatistical nonstationarity can be recognized in a variety of common forms and is often complicated by the scale dependence of spatial nonstationarity (Pyrcz and Deutsch, 2014). Spatial statistics may appear stationary locally but may be nonstationary over a larger spatial intervals, due to long range spatial continuity relative to the area of interest (type I nonstationarity). Another common pattern of spatial nonstationarity results from large-scale variations in the statistics of interest that may be decomposed as additive components of nonstationary trend and stationary residual (type II nonstationarity) (Cuba et al., 2012; Meul and Van Meirvenne, 2003).

Nonstationarity may exist over temporal and spatial intervals, but we focus on spatial nonstationarity. Various spatial modeling approaches are available to account for spatial nonstationarity. Detrending is a common method to decompose the original data into nonstationary trend and stationary residual components and then to model the stationary residuals with statistical and geostatistical modeling (Myers, 1989; Vieira et al., 2010; Wu et al., 2007). Dividing the model into stationary subdomains with statistics of interest modeled independently within each subdomain and then merger of the subdomains is another alternative to account for spatial nonstationarity (Pyrcz and Deutsch, 2014).

There has been a rapid expansion of deep learning methods and workflows for subsurface geostatistical modeling. Mosser et al. (2017) apply generative adversarial neural networks (GANs) to reconstruct the representative samples of three-dimensional porous media at different scales in a more computational efficient way than classical stochastic methods of image reconstruction. Zhang et al. (2019) propose improved integration of geological patterns for conditional 3D nonstationary subsurface reservoir models with GANs with results that exceeded conventional multiple-point simulation. Jo et al. (2020) model realistic 3D lob reservoirs that conserve complex geological features in rule-based models with a GAN. Pan et al. (2022) propose a workflow for multi-scale lobe modeling from lobe scale to lobe element scale, with hierarchical stochastic pix2pix coupled with an automatic segmentation method.

Yet the spatial, subsurface setting continues to pose unique modeling challenges due to ubiquitous spatial nonstationarity. Liu et al. (2022) demonstrate and quantify the impact of spatial nonstationarity on prediction performance of the convolutional neural network (CNN), a common deep learning method for prediction, regression, and classifications from image data. Liu et al. (2022) train a CNN model using stationary images (realizations calculated with the geostatistical sequential Gaussian simulation method) and explore the prediction accuracy over different types of spatial nonstationarity. Their results show that spatial nonstationarity decreases the model prediction accuracy proportional to the degree of nonstationarity.

The transformer, initially proposed for dealing with machine translation tasks in 2007 by Google, is an encoder-decoder architecture that solely depends on the self-attention module to address dependencies in the input and output sequences (Vaswani et al., 2017). The transformer is applied broadly with promising results over other state-of-art deep learning methods, such as CNN in computer vision (CV) tasks (Dosovitskiy et al., 2021; Liu et al., 2021) and recurrent neural network (RNN) in natural language processing (NLP) tasks (Karita et al., 2019; Vaswani et al., 2017). Unlike the convolution process in CNN and the recurrent process in RNN, transformer uses two core mechanisms, the self-attention mechanism and positional embedding, to realize features' extraction.

The self-attention mechanism maps sets of queries and keys to the output, a weighted sum of attention values. This mechanism aims to calculate the representation of a sequence by focusing on the input sequence differentially. The commonly-used method to calculate attention values is scaled dot-product attention (Equation 1) (Vaswani et al., 2017).

$$Attention(Q, K, V) = softmax\left(\frac{QK^T}{\sqrt{d_k}}\right)V \qquad (1)$$

A set of queries, keys, and values are packed together into Q, K, and V matrix respectively. Queries, keys, and values are projection vectors based on the input vector. Q and K both have dimensions of $d_k$, V has the dimension of $d_v$. The attention value is calculated by applying a dot product between Q and K, scaling with $\sqrt{d_k}$, and applying a softmax function on V.

Based on the single attention module, multi-head self-attention (MSA) is proposed to incorporate different parts of representations of input vectors by performing multiple single attentions (Equation 1) in parallel and then concatenating the corresponding attention values with the linear projection weight matrix W (Equation 2) (Vaswani et al., 2017).

$$MultiheadAttention(Q, K, V) = Concat(head_1, \ldots, head_h)W$$
$$where\ head_i = Attention(QW_i^Q, KW_i^K, VW_i^V) \qquad (2)$$

where $W \in \mathbb{R}^{hd_v \times d_{model}}, W_i^Q \in \mathbb{R}^{d_{model} \times d_k}, W_i^K \in \mathbb{R}^{d_{model} \times d_k}$, and $W_i^V \in \mathbb{R}^{d_{model} \times d_v}$, and the target model dimension is $d_{model}$. Apart from the attention module in transformer architecture, another module is the positional encoding that adds positional information to the encoder embedding for the model to learn the order of the input sequence directly instead of indirectly through recurrence and convolution processes (Zhou et al., 2020). Positional embedding methods are generally categorized as absolute positional embedding (APE) and relative positional embedding (RPE). The APE assigns a fixed-value vector to each position in advance (Vaswani et al., 2017), while the RPE estimates bias weights based on the distance between sequence elements to incorporate relative positional information (Ke et al., 2021; Shaw et al., 2018). One common function for APE is the sine and cosine function (Equation 3) (Vaswani et al., 2017)

$$PE_{pos,j} = \begin{cases} \sin(pos/10000^{j/d_m}), j \text{ is even} \\ \cos(pos/10000^{(j-1)/d_m}), j \text{ is odd} \end{cases} \quad (3)$$

where pos represents the position, and $j$ refers to the $j^{th}$ dimension of the embedding vector ($d_m$). Inspired by the successful applications of transformer in the NLP domain, the vision transformer (ViT) architecture is developed to solve CV tasks. ViT applies a standard transformer by treating image patches as words. ViT attains accurate results on several public image classification task benchmarks compared with state-of-the-art CNN architectures (such as BiT, ResNet) (Dosovitskiy et al., 2021), along with panoptic segmentation (Wang et al., 2020), object detection (Carion et al., 2020), and image generation (Zhang et al., 2019).

However, the original self-attention mechanism in transformers has high complexity and limited local attention sensitivity for high-resolution data, and it further has quadratic computational complexity ($O(n^2)$) relative to image size $n$ (Yang et al., 2021). More self-attention variants are developed to improve the flexibility and generalization of transformers for the field of CV. For example, Swin Transformer (SwinT) calculates attentions within local non-overlapping windows that partition the image and bridge the attentions in the proceeding layer through a shifted window process. This shifted window design reduces the computational complexity to be linear ($O(n)$) relative to the image size by limiting the attention calculation in local windows (Liu et al., 2021). Succeeding works build further on SwinT by incorporating local attention and global attention (Chu et al., 2021; Tu et al., 2022), multi-scale attention (Xu et al., 2022), and advanced training strategies (Touvron et al., 2021). Chen et al. (2022) apply self-attention-based

GAN to subsurface modeling of stationary channels and nonstationary delta faces. Their work demonstrates that self-attention-based GANs improve geological structure reproducibility, including in the reproduction of long-distance dependence than traditional GANs, especially in the presence of nonstationarity. Their work proves the applicability of self-attention networks for generative tasks, however, to the best of our knowledge, there is no work applying self-attention networks for predictive tasks in the face of spatial nonstationarity.

We demonstrate and quantify the improvement in spatial subsurface model reproduction in the presence of the common forms of subsurface nonstationarity with vision transformers and benchmark the results against the performance of CNN predictions from Liu et al. (2022). We also propose and quantify the improvement of two novel training techniques for enhancing the transformer performance for nonstationary subsurface modeling, data augmentation and transfer learning.

The methodology section introduces the benchmarking workflow, including the detailed testing model architectures, and then the two proposed training techniques for transformer architecture. The results and discussion sections include summary quantification of the training performance, accuracy comparisons before and after applying training techniques, and prediction performance in the face of various levels of spatial nonstationarity. The conclusions section includes guidance on the implementation of our proposed method, and general guidance for deep learning model choices for spatial nonstationarity, and limitations.

## 2. Methodology

To demonstrate and quantify the improvement in spatial subsurface model reproduction in the presence of geostatistical, spatial nonstationarity with vision transformers we apply the following workflow.

1. Train each alternative deep learning architectures (CNN, ViT, and SwinT) with stationary geostatistical feature realizations of the porosity calculated with sequential Gaussian simulation method to predict the variogram range.
2. Apply data augmentation and transfer learning training techniques to improve the transformer model training performance without modifying the model complexity of the original architectures.
3. Test the model performance with Type I & II nonstationary realizations.

4. Demonstrate and quantify the improvement of deep learning prediction performance in the face of two types of nonstationarity with transformer model compared with CNN.

2.1 Workflow and Training Model Architectures

We apply the workflow from Liu et al. (2022), as illustrated in Figure 1. First, we model a series of unconditional stationary geostatistical porosity feature realizations with variable variogram ranges with sequential Gaussian simulation (SGS) that calculates equal probable realizations using input data, univariate distribution and variogram model. Then we train each deep learning models (CNN, ViT, and SwinT) with these stationary realizations to predict the variogram range. Lastly, we evaluate the prediction performance of well-trained models on the two types of spatial nonstationarity. We compute type I nonstationary realizations (spatial features relatively larger than the domain size) with variogram ranges larger than 1/3 of the domain size. The type II nonstationary realization (spatial trend) is realized by adding a linear trend model to the stationary residual.

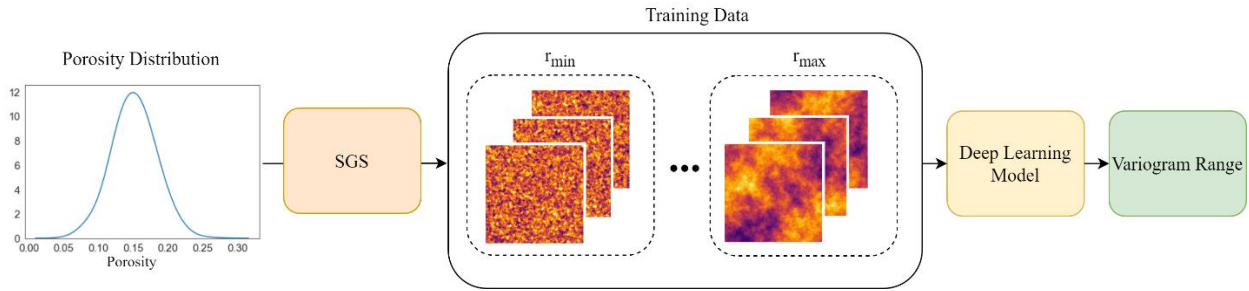

Figure 1. The benchmarking workflow for image generation, and training and testing the deep learning model for the prediction of variogram range. The training data is a series of unconditional stationary realizations of porosity calculated with SGS with variable variogram range from $r_{min}$ to $r_{max}$. The deep learning models are trained to predict the variogram range from the labeled training data.

The training data are stationary realizations with variogram ranges varying from 40 to 400 m (Figure 2, part a). The realization grid size is 224 x 224 cells with a regular grid cell size of 5 x 5 m. We compute 50 realizations for each variogram range by varying the random number seed number to evaluate expected prediction accuracy. Realizations with variogram ranges of 400 to 1000 m are calculated for the type I nonstationarity, an example shown in Figure 2, part b. The variogram ranges for type II nonstationarity are from 40 m to 400 m with trend variance proportions (the proportion of variance described by the trend) from 0% (stationarity only) to 90%, an example presented in Figure 2, part c. 50 realizations are computed for each pair of trend variance and stationary residual variogram range.

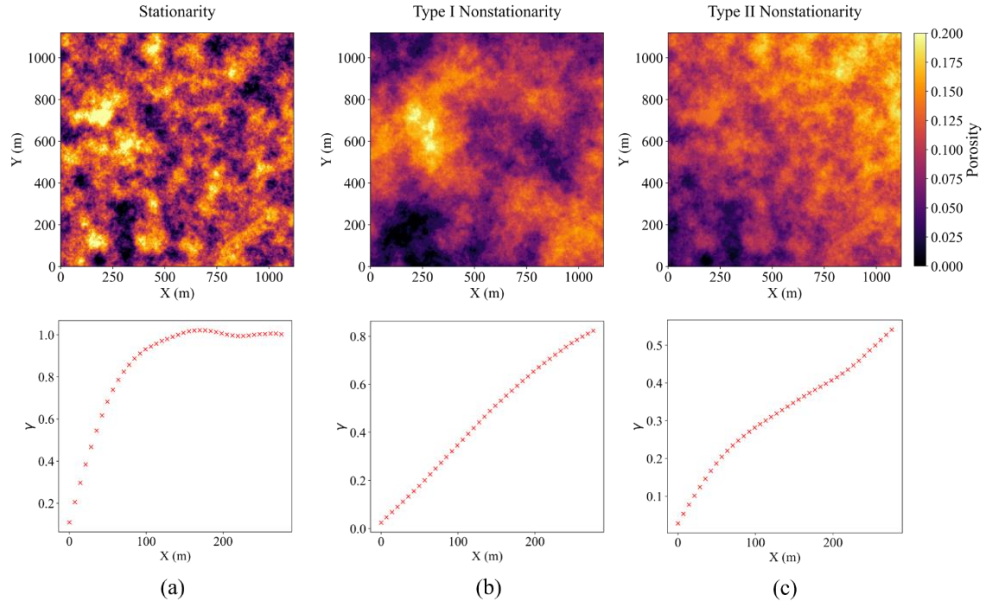

Figure 2. SGS realizations for (a) a stationary case with a variogram range of 100 m, (b) a type I nonstationary case with a variogram range of 600 m, (c) a type II nonstationary case with a linear trend describing 80% of the variance and residual variogram range of 100 m.

We train CNN, ViT, and SwinT models to predict the variogram range from labeled SGS realizations. The CNN architecture (Figure 3) maps the input image of size 224 x 224 x 1 to the response feature, variogram range, through 3 convolutional layers with ReLU activation function, followed by a pooling layer, a convolutional layer and a pooling layer, and a fully connected layer. The ViT architecture (Figure 4, part a) follows the steps: (1) image patch partition to same-sized patches with a patch size of 28 x 28 grids; (2) linear projection of flattened patches and combination of positional embedding; (3) 6 layers of transformer encoder to calculate multi-head self-attention values (Figure 4, part b); (4) the multi-layer perceptron (MLP) regression head outputs the target label (Dosovitskiy et al., 2021). The SwinT architecture (Figure 5, part a) consists of 4 stages after the image patch partition, dividing the size H x W image into patches of size 4 x 4 grids. The Swin Transformer block in each stage has two successive layers ($z^l$ and $z^{l+1}$), one is for MSA calculation within each window (W-MSA), and another one is for MSA calculation through shifted windows

(SW-MSA) (Liu et al., 2021). We design all three models to have similar model complexities for a fair comparison (Steiner et al., 2022): 240 k model parameters for CNN, 226 k model parameters for ViT, and 233K model parameters for SwinT.

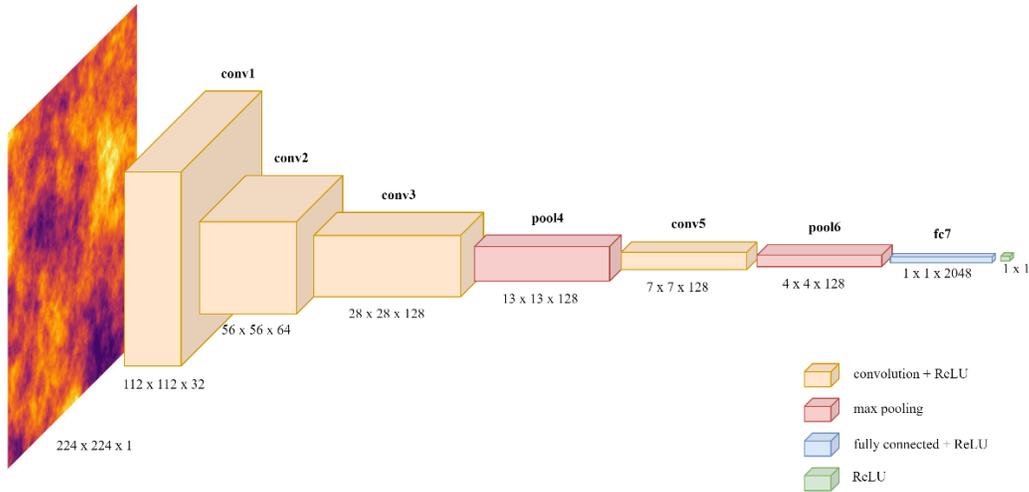

Figure 3. the CNN architecture applied to predict variogram range from SGS images.

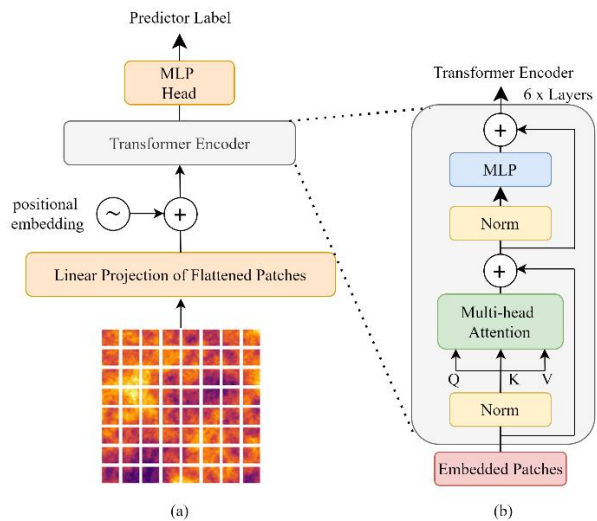

Figure 4. the ViT architecture, (a) the main architecture (b) the transformer encoder module to predict variogram range from SGS images.

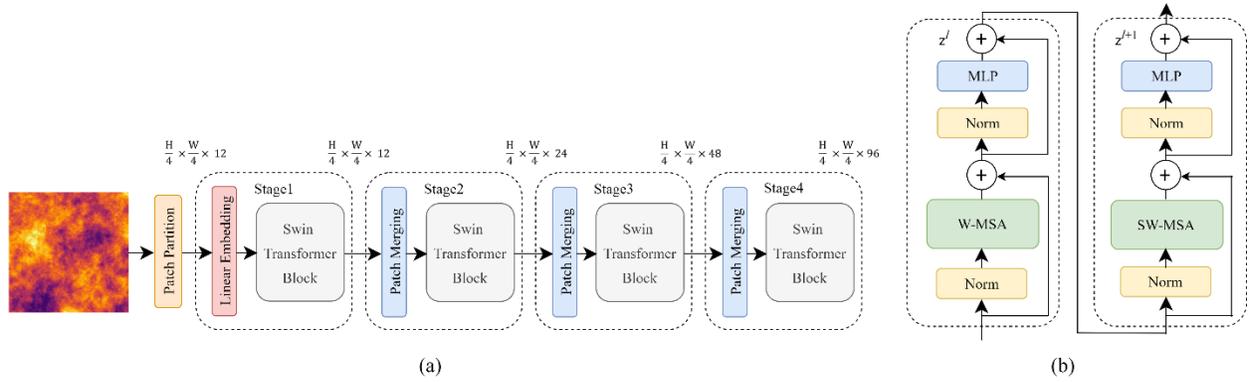

Figure 5. the SwinT architecture (a) the main architecture (b) Swin Transformer Blocks (adapted from Liu et al., (2021)) to predict variogram range from SGS images.

## 2.2 Data Augmentation

Transformer models require a large number of training data (Dosovitskiy et al., 2021; Ho et al., 2019; Liu et al., 2021; Liu et al., 2022; Touvron et al., 2021). Data augmentation is commonly applied before training transformer models to satisfy enhanced data demands and improve model performance (Steiner et al., 2022). Image data augmentation focuses on enhancing the size and quality of training datasets and can be realized by simple image transformations, such as flipping, rotating, and random cropping (Shorten and Khoshgoftaar, 2019). Compared to CNN, vision transformers can be more reliant on data augmentation with smaller training datasets (Steiner et al., 2022). To evaluate the data augmentation on improving transformer model performance, we apply flipping and rotating to increase the training data from 3 k to 10 k and train three architectures again, then compare the training performance with the base case (2.1), with only 3 k realizations of training data.

## 2.3 Transfer Learning

Transfer learning is another training technique for improving training performance and reducing computational complexity by transferring the knowledge learned from previous tasks to downstream tasks (Pan and Yang, 2010). This technique applies the weights of pre-trained models on general, data-rich tasks (e.g., ImageNet) before training target tasks (Raffel et al., 2020; Steiner et al., 2022; Touvron et al., 2021). To demonstrate the potential for reduced computational complexity, we apply transfer learning along with data augmentation. The base case model parameters are initialized with random weights and the comparison group models are initialized with pre-trained weights from the ImageNet dataset (Wightman, 2019). All the model architecture designs are held constant, e.g., embedding dimension, depth (number of layers), and number of heads.

## 3. Results and Discussion

This section presents the testing performance of the three models, CNN, ViT, and SwinT, for the baseline case that does not use data augmentation and the comparison test with data augmentation and transfer learning techniques (3.1). Then we compare two CNN and SwinT models with type I and II nonstationary data and the corresponding results are shown in 3.2 and 3.3.

### 3.1 Three Models' (CNN, ViT, and SwinT) Prediction Performance

The base case CNN, ViT and SwinT models are fully trained with 3k stationary porosity realizations. The same architectures for the ViT and SwinT models are trained, initialized with random model parameters, again using data augmentation techniques (resulting in 10 k training images) and with the transfer learning technique. We test the performance of all the fully-trained models with another 150 stationary realizations and their prediction errors distributions are shown in Figure 6. The prediction error per realization is calculated by $\frac{y-\hat{y}}{y}$ where $y$ is the true variogram range and $\hat{y}$ is the predicted variogram range. We only apply transfer learning techniques to train transformer models rather than CNN model due to the unavailability of open source public pre-trained CNN model. The data augmentation technique effectively improves the transformer model's performance, especially for SwinT (Figure 6, case 2). The ViT improves further with transfer learning, reaching the similar performance with CNN and SwinT (Figure 6, case 3). The pre-trained weights of the transfer learning technique are from the original ViT model trained with ImageNet-1k and fine-tuned from ImageNet-21k (Dosovitskiy et al., 2021; Wightman, 2019) and the SwinT model is pre-trained with ImageNet-22k and fine-tuned with ImageNet-1k (Wightman, 2019). From the three sets of testing performance comparisons, we choose SwinT as our competitive transformer model and CNN as our base model to demonstrate the improvements of deep learning model prediction performance in the face of type I and II nonstationarity.

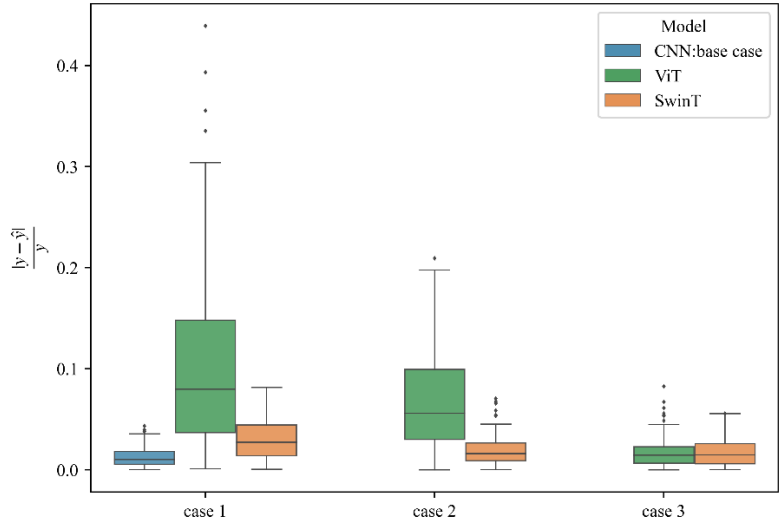

Figure 6. the box plots of CNN, SwinT, and ViT models prediction errors for stationary realizations of the case 1 (no data augmentation nor transfer learning), case 2 (data augmentation), and case 3 (transfer learning). The prediction error is calculated by $\frac{y-\hat{y}}{y}$ where y is the true variogram range and $\hat{y}$ is the predicted variogram range.

### 3.2 Type I Nonstationarity

We compare the CNN and SwinT model performance with the type I nonstationary realizations of the porosity over 20 realizations for each variogram range for the steady statistic calculation of prediction errors. Figure 7 shows the box plots of relative prediction errors, $\frac{y-\hat{y}}{y}$. Overall, both models show an increasing trend of prediction errors with increasing variogram ranges (higher degrees of type I nonstationarity). However, we observe that SwinT improves the prediction performance in the presence of the type I nonstationarity compared with CNN, and such improvement increases with increasing variogram ranges, e.g., the improvement of prediction error is about 10% at the variogram range of 500 m and more than 20% at the variogram range of 800 m.

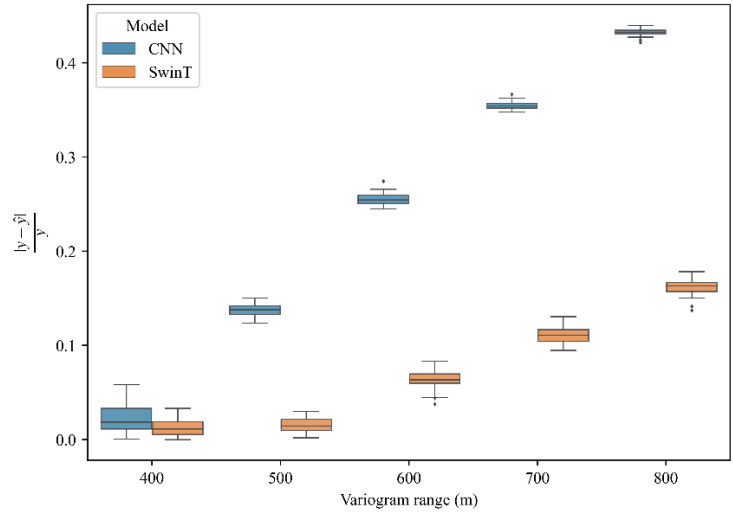

Figure 7. CNN and SwinT prediction performance (box plots of prediction errors) comparison in the presence of Type I nonstationarity.

## 3.3 Type II Nonstationarity

We calculate combinatorial of type II nonstationary realizations of the porosity with various variogram ranges from 40m to 400m and different trend variance proportions (the proportion of variance described by the trend) from 0% (stationarity only) to 90%. For each pair of variogram range and trend variance proportion, 20 realizations are applied for the calculation of the expected prediction errors. Figure 8 (part a) shows the mean of relative errors of prediction of CNN model, and (part b) for the SwinT model. SwinT improves the prediction performance over the moderate variogram ranges with larger trend variance proportions, for example the realizations with pairs of variogram ranges from 100 m to 300 m and trend variance proportions above 40% roughly. However, at a very small variogram range, for example 40 m, SwinT does not outperform CNN. One possible explanation is that the self-attention calculation is based on the whole input data which is more appropriate for characterizing the large-scale features, while the CNN feature calculation is realized through the filter convolution which characterizes features over the scale of the composite of the convolution kernels from each layer.

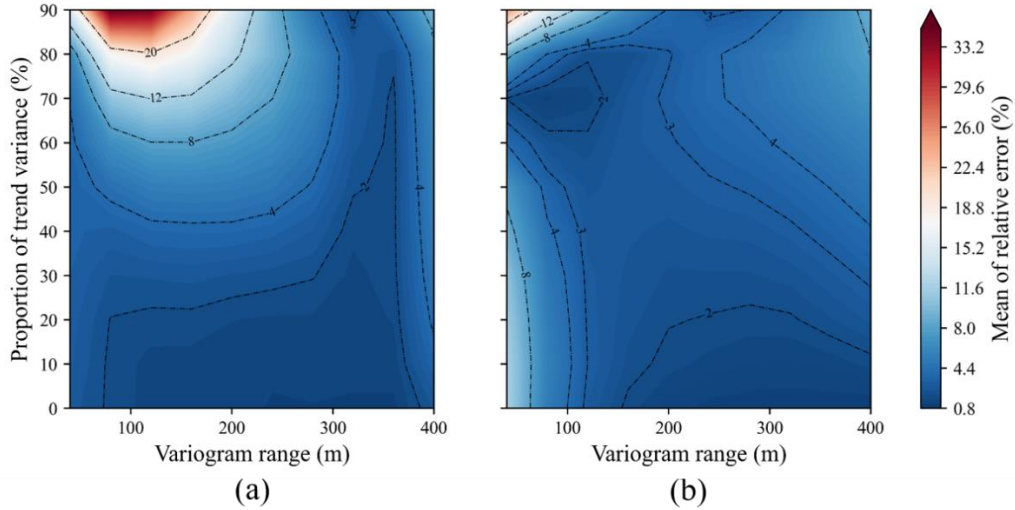

Figure 8. the prediction performance (contour plots of mean of relative errors) comparison in the presence of Type II nonstationarity of (a) CNN, and (b) SwinT.

## 4. Conclusions

We demonstrate and evaluate the improvement of deep learning model in predicting the variogram range in the face of typical two types of geostatistical spatial nonstationarity using vision transformer models. Our result shows that transformer models may mitigate the impact of geostatistical spatial nonstationarity on deep learning model prediction performance. For the type I nonstationarity, the prediction improvement by transformer models increases with increasing variogram ranges. Regarding the type II nonstationarity, the transformer models effectively improve the prediction performance, especially for moderate variogram ranges with high trend variance proportions. Our work proves that the transformer model is more appropriate for modeling large-scale relationships while mitigating nonstationarity. By demonstrating the deep learning model prediction performance improvement in spatial nonstationarity, we hope this work provides guidance for deep learning applications for nonstationary datasets commonly encountered in natural settings.

## 5. Acknowledgments

The authors thank the support from the Digital Reservoir Characterization Technology (DIRECT) Industry Affiliate Program at the University of Texas at Austin. The unclassified release number for the work is LA-UR-22-32653.